\pdfoutput=1

\documentclass[11pt]{article}

\usepackage[final]{coling}

\usepackage{times}
\usepackage{latexsym}

\usepackage[T1]{fontenc}

\usepackage[utf8]{inputenc}

\usepackage{microtype}

\usepackage{inconsolata}

\usepackage{graphicx}

\usepackage{enumitem}
\setenumerate[1]{itemsep=0pt,partopsep=0pt,parsep=\parskip,topsep=5pt}
\setitemize[1]{itemsep=0pt,partopsep=0pt,parsep=\parskip,topsep=5pt}
\setdescription{itemsep=0pt,partopsep=0pt,parsep=\parskip,topsep=5pt}

\title{Tutorial Proposal:\\Speculative Decoding for Efficient LLM Inference}

\author{{Heming Xia}\textsuperscript{$\dagger$}, {Cunxiao Du}\textsuperscript{\rm $\ddagger$}, {\textbf{Yongqi Li}}\textsuperscript{\rm $\dagger$}, {Qian Liu}\textsuperscript{\rm $\ddagger$}, {\textbf{Wenjie Li}}\textsuperscript{\rm $\dagger$}\\
  \textsuperscript{\rm $\dagger$}The Hong Kong Polytechnic University,
  \textsuperscript{\rm $\ddagger$}Sea AI Lab\ \\
  {\tt \{he-ming.xia\}@connect.polyu.hk; \{cnsdunm\}@gmail.com}
}

\begin{document}
\maketitle
\begin{abstract}
This tutorial presents a comprehensive introduction to Speculative Decoding (SD), an advanced technique for LLM inference acceleration that has garnered significant research interest in recent years. SD is introduced as an innovative decoding paradigm to mitigate the high inference latency stemming from autoregressive decoding in LLMs. At each decoding step, SD efficiently drafts several future tokens and then verifies them in parallel. This approach, unlike traditional autoregressive decoding, facilitates the simultaneous decoding of multiple tokens per step, thereby achieving promising 2$\times$$\sim$4$\times$ speedups in LLM inference while maintaining original distributions. This tutorial delves into the latest techniques in SD, including draft model architectures and verification strategies. Additionally, it explores the acceleration potential and future research directions in this promising field. We aim for this tutorial to elucidate the current research landscape and offer insights for researchers interested in Speculative Decoding, ultimately contributing to more efficient LLM inference.
\end{abstract}

\section{Introduction}
Large Language Models (LLMs) have achieved significant progress across various domains, e.g., machine translation, fact verification, and conversational systems~\citep{openai:2023gpt4, Hugo:2023llama, Hugo:2023llama2}. However, the token-by-token generation nature necessitated by autoregressive decoding falls short in efficiency, especially as model sizes increase~\cite{Pope:2023efficiency}. The inference latency primarily stems from the memory-bound computation of LLMs~\cite{Patterson:2004latencybandwith, Shazeer:2019memorybandwith}. Each forward pass requires transferring all LLM parameters from High-Bandwidth Memory (HBM) to the on-chip cache of modern accelerators like GPUs. This process, which produces only one token per step, leads to the underutilization of these accelerators and results in inefficiencies.

To mitigate the high inference latency, Speculative Decoding (SD) has been introduced as an innovative decoding paradigm~\cite{Stern:2018blockwise, Chen:2023specsampling, xia:2022specdec, Leviathan:2023specdec}. The key idea of SD is to exploit the token-level computation parallelism of LLMs by integrating an efficient ``draft'' model into the inference process. At each decoding step, SD first utilizes this draft model to efficiently predict multiple future tokens and then validates all these tokens with the target LLM in parallel~\cite{xia2024unlocking}. Since the draft model requires much fewer computation resources and the cost of parallel verification is nearly equivalent to decoding a single token, SD achieves considerable speedups by substantially reducing overall decoding steps, which diminishes the need for frequent memory operations of LLM parameters~\cite{medusa, Li:2024eagle, Fu:2023lookahead, Du:2024glide}. Furthermore, SD has been theoretically proven to maintain identical distributions to the target LLM~\cite{Leviathan:2023specdec, Chen:2023specsampling}, which broadens its application in both academic and industrial communities.

The primary factors for the acceleration effect of SD are threefold: \textbf{1)} the inference efficiency of the introduced draft model, \textbf{2)} the acceptance rate of the drafted tokens, and \textbf{3)} the average token acceptance length throughout the inference process. To fully exploit the inference parallelism and obtain high speedups, the structure of the draft model, the drafting mechanism, as well as the verification strategy of LLMs play a vital role in SD. In this tutorial, we will present a comprehensive introduction to this innovative decoding paradigm. We will start by providing preliminaries covering the foundational concepts of LLMs (e.g., decoder-only LLMs and autoregressive decoding), and illustrating the memory bottleneck in LLM inference. We will then focus on recent advancements in \textit{drafting and verification strategies}, \textit{leading algorithms}, and \textit{applications} of Speculative Decoding.

\paragraph{A taxonomy of methods} We introduce a taxonomy of SD methods based on a variety of dimensions. SD approaches can be categorized by \textit{draft model architectures} into two main types: 1) independent drafting methods such as leveraging smaller LMs from the same model series~\cite{Leviathan:2023specdec, Chen:2023specsampling, Miao:2023specinfer, Zhou:2023distillspec, He:2023REST} and 2) self-drafting approaches that integrate lightweight draft modules into the target LLM~\cite{medusa, Li:2024eagle, Du:2024glide, Fu:2023lookahead}. At the same time, SD methods can be categorized based on \textit{the supported verification strategy}, such as the greedy decoding~\cite{Stern:2018blockwise, xia:2022specdec, Zhang:2023draftverify}, speculative sampling~\cite{Leviathan:2023specdec, Chen:2023specsampling}, and token tree verification~\cite{Miao:2023specinfer, He:2023REST, Li:2024eagle, li:2024eagle2}, an effective approach to increase the token acceptance rate.

\paragraph{Cutting-edge algorithms} Then, we offer a thorough discussion and analysis of cutting-edge SD approaches. Specifically, we delve into an in-depth analysis of the state-of-the-art methods, Eagle~\cite{Li:2024eagle} and Eagle-2\cite{li:2024eagle2}, which achieve an average 4$\times$ speedup across various generation tasks. This analysis covers their drafting model architectures, drafting mechanisms, and innovative verification strategies. Furthermore, we conduct discussions on other leading methods such as Medusa~\cite{medusa}, GliDe with a CaPE~\cite{Du:2024glide}, and Lookahead Decoding~\cite{Fu:2023lookahead}, offering insights for researchers interested in this promising research area.

\paragraph{Evaluations and applications} After discussing the fundamental components and prominent variants of Speculative Decoding, we present a fair comparison of various open-source SD methods using a unified testing environment and evaluation platform.  This comparison aims to raise awareness among academics about the practical speedups expected from SD. Beyond general generation tasks, we further introduce SD techniques in downstream applications, such as retrieval-augmented SD~\cite{Zhang2023RaLMSpec, wang2024:specrag}, long-context SD~\cite{sun2024:triforce}, and multimodal SD~\cite{gagrani2024:mmspec}. Additionally, we discuss SD methods tailored for specific input contexts~\cite{Sun:2020SAD, yang:2023llma, saxena2023pld} and advanced SD approaches for mobile phone applications~\cite{Xu:2023LLMCad}.

Finally, we will demonstrate the effectiveness of Speculative Decoding through a practical exercise. We conclude this tutorial by summarizing the strengths and challenges of SD and discussing several important future directions. These include: 1) how we can further improve the speedups of SD methods by optimizing the trade-offs between speculation accuracy and drafting efficiency, 2) how to apply Speculative Decoding in batched inference scenarios, and 3) how to integrate Speculative Decoding with other leading techniques. We hope this tutorial serves as an essential guide for newcomers to this field and motivates further research.

\section{Target Audience}
This tutorial will be accessible to anyone with a basic knowledge of machine learning and natural language processing. We believe the topic will be of interest to both NLP researchers and students in academia, as well as NLP practitioners in the industry, particularly those interested in LLM efficiency, sparsity, and computational parallelism. By providing a systematic overview of recent promising approaches for this rapidly evolving field, this tutorial hopefully reveals new research opportunities to the audience.

\section{Outline}
\paragraph{1. Introduction} (15 minutes)
\begin{itemize}
    \item An overview of the tutorial
    \item The benefits of Speculative Decoding
\end{itemize}

\paragraph{2. Preliminaries} (15 minutes)
\begin{itemize}
    \item Auto-regressive decoding
    \item Memory bottleneck in LLM inference
\end{itemize}

\paragraph{3. Speculative Decoding: A taxonomy of methods} (40 minutes)
\begin{itemize}
    \item Definition, formulation, and illustrated algorithms of Speculative Decoding
    \item Introduction to various draft model architectures and verification strategies
\end{itemize}

\paragraph{4. Speculative Decoding: Cutting-edge algorithms} (40 minutes)
\begin{itemize}
    \item Discussions and analysis of the state-of-the-art algorithms
    \item Advanced techniques in Speculative Decoding: adaptive token tree verification, draft model alignment, and etc
\end{itemize}

\paragraph{5. Speculative Decoding: Metrics and evaluations} (20 minutes)
\begin{itemize}
    \item Evaluation metrics of Speculative Decoding
    \item Fair comparisons and discussions of leading methods across diverse scenarios
\end{itemize}

\paragraph{6. Speculative Decoding: Downstream adaptations} (20 minutes)
\begin{itemize}
    \item Retrieval-augmented Speculative Decoding
    \item Long-context and multimodal Speculative Decoding
\end{itemize}

\paragraph{7. Demonstration: An exercise to show how Speculative Decoding works} (10 minutes)

\paragraph{8. Conclusions and future directions} (10 minutes)

\section{Diversity Considerations}

The speakers are from two academic institutions with an affiliation and an academic research group, including both a professor and Ph.D. students. The methods presented in our tutorials are language-agnostic and can be extended to non-English contexts and we also offer a brief overview of several papers focusing on multilingual and expert-domain extensions of the core frameworks. We will reach out to both academic and industry communities such as CAMEL-AI\footnote{\url{https://www.camel-ai.org/}} and DeepSeek\footnote{\url{https://www.deepseek.com/}}, to encourage diverse audience participation in our tutorial. Given that speculative decoding effectively accelerates LLM inference while maintaining unchanged distributions, we anticipate this tutorial will be particularly beneficial for researchers with modest computational resources who may not have access to extensive hardware infrastructure.

\section{Other Information}

\paragraph{Type of the tutorial} 
Cutting-edge.

\paragraph{Length}
This is a 4-hour tutorial.

\paragraph{Breath} 
We estimate that approximately 20\% of the tutorial will center around work done by the presenters. The papers we will cover are from both academia and industry.

\paragraph{An estimate of the audience size} 
Given that Speculative Decoding is now applied in a various range of NLP platforms such as vLLM~\cite{Kwon:2023vLLM}, Transformers~\cite{wolf:2020huggingface}, and PyTorch~\cite{pytorch} to accelerate LLM inference, we estimate that the number of audiences will be around 100+.

\paragraph{Technical equipment.} We would like to have
Internet access to show online demos.

\paragraph{Open access} We plan to make all teaching material available online and agree to allow the publication of slides and video recordings in the COLING anthology.

\paragraph{Pedagogical material} We plan to do some short hands-on exercises to let the audience try different Speculative Decoding techniques to accelerate LLM inference using Colab.

\section{Presenters}
\paragraph{Heming Xia}
Heming Xia is a Ph.D. student at the Natural Language Processing Group of The Hong Kong Polytechnic University, advised by Prof. Wenjie Li. His research interest lies in natural language processing and machine learning. His recent research focuses on speculative decoding, tool learning, vision-language understanding, and representation learning of LMs. He has served as the reviewer for top-tier conferences like ACL, EMNLP, and Neurips. He was recognized as a Merit Student at Peking University in 2021.

\paragraph{Cunxiao Du}
Cunxiao Du is a Research Scientist in the Sea AI Lab. He received a PhD degree from The Singapore Management University and a Bachelor's degree from Shandong University. His research interests include LLM, especially, the efficiency of LLM and parallel decoding of LLM. He has also served as the reviewer for top-tier conferences like ICML, ICLR, and Neurips. 

\paragraph{Yongqi Li}
Yongqi Li is a Postdoctoral Fellow in the Department of Computing at The Hong Kong Polytechnic University. He received a PhD degree from The Hong Kong Polytechnic University and a Bachelor's degree from Shandong University. His research interests include natural language processing, information retrieval, and multimedia computing. He has published about 20 papers in leading conferences and journals, such as ACL, CVPR, SIGIR, KDD, AAAI, CIKM, TOIS, and IPM. He has also served as an area chair for COLING 2025, a guest editor for Frontiers in Big Data, and a reviewer for TKDE, TOIS, TMM, and TCVST journals.

\paragraph{Qian Liu}
Qian Liu is a Research Scientist at Sea AI Lab, Singapore. Before he joined Sea AI Lab, he was a joint Ph.D. candidate at Beihang University and Microsoft Research Asia. His research interests include code generation and reasoning. He has published more than 20 papers in top-tier conferences, and his representative works include TAPEX, LoraHub, and StarCoder. He was nominated for the Baidu Scholarship 2020 and the KAUST AI Rising Star 2024.

\paragraph{Wenjie Li}
Wenjie Li is a Professor in the Department of Computing at the Hong Kong Polytechnic
University, Hong Kong. She received a Ph.D. degree in systems engineering and engineering management from the Chinese University of Hong Kong, Hong Kong, in 1997. Her main research interests include natural language understanding and generation, machine conversation, summarization, and question answering. Wenjie served as the program chair for ACL 2021 and (senior) area chair for many *ACL conferences. 

\section{Reading List}
\begin{itemize}
    \item Unlocking Efficiency in Large Language Model Inference: A Comprehensive Survey of Speculative Decoding~\cite{xia2024unlocking}
    \item Blockwise Parallel Decoding for Deep Autoregressive Models~\cite{Stern:2018blockwise}
    \item Speculative Decoding: Exploiting Speculative Execution for Accelerating Seq2seq Generation~\cite{xia:2022specdec}
    \item Fast Inference from Transformers via Speculative Decoding~\cite{Leviathan:2023specdec}
    \item Accelerating Large Language Model Decoding with Speculative Sampling~\cite{Chen:2023specsampling}
    \item Speculative Decoding with Big Little Decoder~\cite{Kim:2023bild}
    \item Inference with Reference: Lossless Acceleration of Large Language Models~\cite{yang:2023llma}
    \item Accelerating Transformer Inference for Translation via Parallel Decoding~\cite{Santilli:2023paralleldecoding}
    \item SpecInfer: Accelerating Generative Large Language Model Serving with Speculative Inference and Token Tree Verification~\cite{Miao:2023specinfer}
    \item Draft \& Verify: Lossless Large Language Model Acceleration via Self-Speculative Decoding~\cite{Zhang:2023draftverify}
    \item REST: Retrieval-Based Speculative Decoding~\cite{He:2023REST}
    \item Medusa: Simple LLM Inference Acceleration Framework with Multiple Decoding Heads~\cite{medusa}
    \item EAGLE: Speculative Sampling Requires Rethinking Feature Uncertainty~\cite{Li:2024eagle}
    \item GliDe with a CaPE: A Low-Hassle Method to Accelerate Speculative Decoding~\cite{Du:2024glide}
    \item Break the Sequential Dependency of LLM Inference Using Lookahead Decoding~\cite{Fu:2023lookahead}
\end{itemize}

\section*{Ethics Statement}
Speculative Decoding (SD) is an effective LLM inference acceleration technique that maintains unchanged distributions of LLMs. Most advanced SD techniques do not require full-parameter re-training, which makes it more energy efficient and can reduce carbon footprints. Previous research also shows that SD can be integrated with other decoding strategies to improve generation quality further. However, we note that SD may copy relevant text spans from historical contexts or retrieve raw data from a corpus, potentially leaking privacy-sensitive information, especially when built on a private corpus. Though most SD methods do not alter the output distributions of LLMs, privacy information may leak through other statistical information, such as overall speedups. We acknowledge this to caution those who manage to apply SD techniques in privacy-sensitive domains.

\bibliography{custom}

\end{document}